\documentclass{esannV2}
\usepackage{bbm}
\usepackage{graphicx}
\usepackage[utf8]{inputenc}
\usepackage{amssymb,amsmath,array}
\usepackage{booktabs}         

\usepackage[table]{xcolor}
\usepackage{colortbl} 
\usepackage{arydshln} 
\usepackage[labelfont=bf]{caption} 
\usepackage{bm}

\definecolor{SecondBg}{RGB}{235,241,255}
\definecolor{BestBg}{RGB}{214,228,255}
\newcommand{\best}[1]{\cellcolor{BestBg}\textbf{#1}}
\newcommand{\second}[1]{\cellcolor{SecondBg}#1}

\begin{document}

\title{Context-Aware Graph Attention for Unsupervised Telco Anomaly Detection}

\author{Sara Malacarne$^{1}$, Eirik Hoel-H\o iseth$^{1}$, Erlend Aune$^{3,5}$,\\ David Zsolt Biro$^2$, Massimiliano Ruocco$^{3,4}$
\thanks{This work was supported by the Norwegian Research Council projects ML4ITS (312062) and SFI NorwAI (309834).}
\vspace{.3cm}\\
1- Telenor Research and Innovation, Norway\\
2- Telenor Denmark OSS \& Tech Analytics, Denmark\\
3- Norwegian University of Science and Technology (NTNU), Norway\\
4- SINTEF Digital, Norway \\
5- Hance, Norway
}

\maketitle

\begin{abstract}
We propose C-MTAD-GAT, an \emph{unsupervised}, \emph{context-aware} graph-attention model for anomaly detection in multivariate time series from mobile networks. C-MTAD-GAT combines graph attention with lightweight context embeddings, and uses a deterministic reconstruction head and multi-step forecaster to produce anomaly scores. Detection thresholds are calibrated \emph{without labels} from validation residuals, keeping the pipeline fully unsupervised. On the public TELCO dataset, C-MTAD-GAT consistently outperforms MTAD-GAT and the Telco-specific DC-VAE, two state-of-the-art baselines, in both event-level and pointwise F1, while triggering substantially fewer alarms. C-MTAD-GAT is also deployed in the Core network of a national mobile operator, demonstrating its resilience in real industrial settings.
\end{abstract}

\section{Background and Motivation}

Mobile networks require anomaly detection (AD) that is fully unsupervised,
computationally efficient, and remains reliable under regime shifts, even when
no labelled incidents are available.
Multivariate Key Performance Indicators (KPIs) are high-dimensional, temporally dependent, and
interdependent, so detectors must couple temporal dynamics with cross–feature
structure while scaling across thousands of \emph{network elements} (NEs),
such as base stations in the Radio Access Network (RAN) or Core functions that
expose KPI time series. To remain deployable in such settings, models must be
compact: in our TELCO configuration, C-MTAD-GAT has about $4.9$M trainable parameters ($\approx10$MB), making it lightweight for continuous deployment in the extensive network infrastructure.

Existing deep unsupervised time-series AD methods typically learn to
\emph{reconstruct} or \emph{forecast} normal behaviour and then flag large
residuals as anomalies, using encoder–decoders~\cite{OmniAnomaly, Hundman_2018, DAGMM}
or GANs~\cite{mad-gan}. Graph-based models such as MTAD-GAT~\cite{mtad-gat}
extend this with temporal and feature-wise attention and a VAE
reconstruction head trained with an ELBO loss.

Within mobile networking, much deployed AD still relies on classical
statistics or standard machine learning methods on aggregated indicators rather than full
multivariate time series, e.g.\ PCA with finite-state machines for
root-cause analysis~\cite{ramirez2023explainable} or feature selection
with Random Forests and SVMs on summary KPIs~\cite{oleiwi2022mlts}. Closer
to our setting, TransKS~\cite{zheng2023transks} and DC-VAE~\cite{garcia2023onemodel}
both use deep models (Transformer or dilated-convolutional VAE) on Telco KPIs
but rely on sensitivity parameters tuned from
labelled anomalies.
By contrast, we propose C-MTAD-GAT, a context-aware extension of
MTAD-GAT~\cite{mtad-gat} designed as a fully unsupervised, \emph{centralised}
detector, that is, for each network domain (e.g.\ RAN or Core) a \emph{single}
graph-attention (GAT) model with temporal and categorical context embeddings is
trained on multivariate time series from thousands of heterogeneous NEs. It
produces per-NE, per-feature anomaly scores with thresholds set by simple
statistics on validation errors, avoiding label-based calibration and making
one compact centralised model much easier to retrain and monitor than per-NE detectors.
 In this work we validate C-MTAD-GAT on the public TELCO dataset \cite{garcia2023onemodel}. The same backbone has been deployed in an operational Core network and piloted in the RAN, achieving comparable performance in both domains.

\textbf{Contributions.}
(i) We introduce C-MTAD-GAT, a \textit{centralised}, \emph{context-aware} extension
of MTAD-GAT that injects static and dynamic metadata via lightweight
context-conditioned convolutions and graph attention, so that a single
model can handle thousands of heterogeneous NEs;
(ii) we instantiate this backbone with a deterministic GRU-based
reconstruction head and a multi-step forecasting head, derive
per-feature anomaly scores from their residuals, and estimate thresholds
from validation errors only, keeping the entire calibration pipeline
label-free; and (iii) we benchmark C-MTAD-GAT against MTAD-GAT and
DC-VAE on the public TELCO dataset using both event-wise affiliation and
pointwise metrics (Macro/Micro/Union).

\section{Problem Setting and Methodology}
\begin{figure}[h]
    \centering
    \includegraphics[width=1\linewidth]{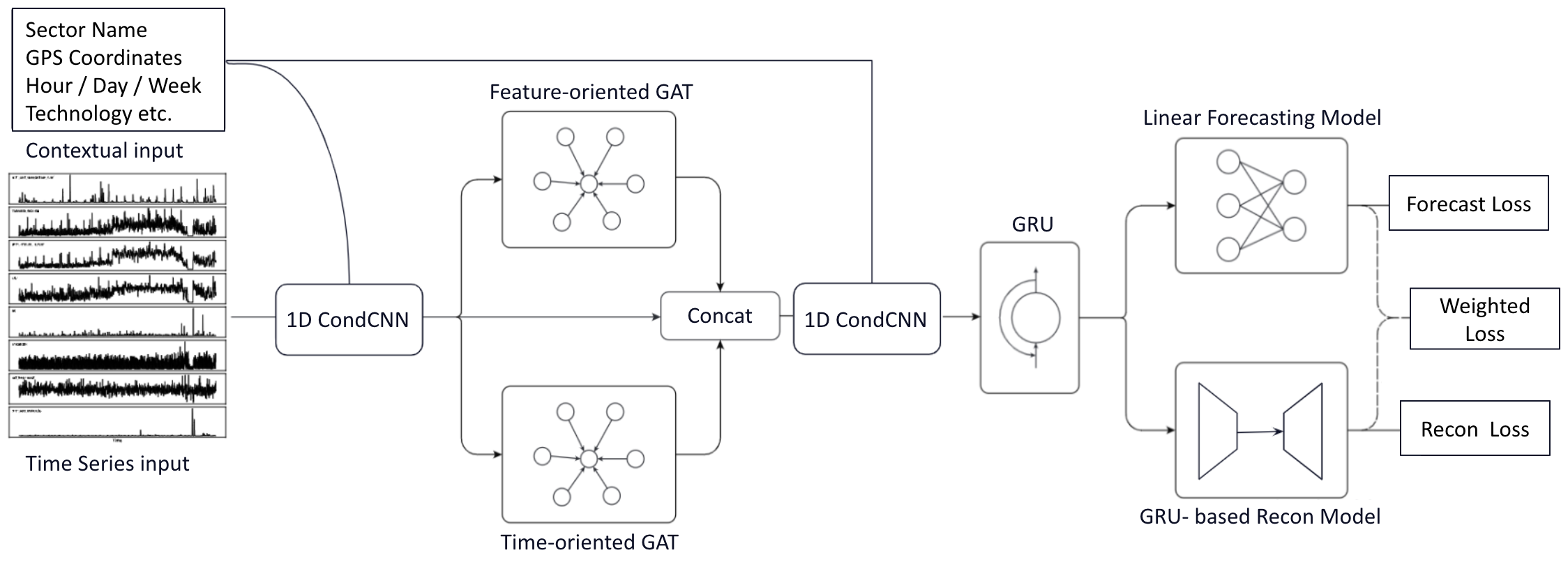}
    \caption{Overview of C-MTAD-GAT: the architecture jointly models temporal dynamics and cross-feature dependencies, while conditioning on contextual input; anomalies are identified by measuring how much each feature deviates from learned reconstructions and forecasts.
    }
    \label{fig:C_MATDGAT}
\end{figure}

We build on MTAD-GAT and specialise it for centralised, context-aware anomaly
detection on telecom KPIs. For each network domain (RAN, Core, TELCO) we train a single model on sliding windows of 
$F$ dynamic real-valued KPIs over 
$W$ timesteps, constructed across many NEs. Each window may also carry
static categorical features (e.g.\ cell type, vendor), dynamic categorical
features (e.g.\ hour-of-day, weekday), and static real-valued metadata
(e.g.\ location).

\textit{Context embedding and conditional convolutions.}
Static and dynamic categorical features are encoded with separate
embedding layers, and static real-valued metadata are projected to the
same space; together they form a per-timestep context representation
$C_t$. In each block we convolve the KPI channels $X_t$, map $C_t$
through a small MLP to the same channel dimension, concatenate the two
along the channel axis, pass them through a two-layer 1D CNN, and add a
skip connection from $X_t$. Disabling context reduces the block to a
standard residual 1D convolution. Conditioning on $C_t$ lets a single
centralised model account for systematic differences across diverse NEs.

\textit{Dynamic feature and temporal attention.}
Analogously to MTAD-GAT~\cite{mtad-gat}, from the initial convolutional
features we compute two attention-enhanced views: a feature-attention
layer that treats KPI dimensions as graph nodes and a temporal-attention
layer that treats timesteps as nodes. We replace the static attention
mechanism with GATv2~\cite{brody2022how}, yielding dynamic attention
scores. 

\textit{Forecasting and reconstruction heads.}
From the encoder state we derive two outputs: (i) a fully-connected
multi-step forecaster that predicts $H$ future steps for all $F$ KPIs in a
single shot, and (ii) a Gated Recurrent Unit- (GRU) based decoder that reconstructs the input window.
Thus, in contrast to MTAD-GAT’s VAE reconstruction head and ELBO loss,
C-MTAD-GAT uses a deterministic GRU-AE without latent sampling
or a KL regulariser. The model is trained end-to-end by minimising the sum
of forecasting and reconstruction losses.

\textit{Anomaly scoring and thresholds.}
At inference we combine the squared forecast and reconstruction residuals
into a single error per timestamp and feature. For each feature we set a
threshold as a constant multiple of its mean validation error and flag an
anomaly whenever the error exceeds this threshold.


\section{Experimental Analysis}
\subsection*{TELCO Dataset}
TELCO is a multivariate time-series dataset from a live production
mobile network~\cite{garcia2023onemodel}. It comprises 12 KPIs,
TS1–TS12, sampled every \emph{5 minutes} over \emph{seven months}
(Jan~1–Jul~31, 2021), with per-series anomalous events manually labelled
by network experts. We follow the original time-ordered split into
train, validation and test sets. Anomalies are rare, so the dataset is
\textit{strongly imbalanced}: on the test period there are $25\,143$
timestamps per series and $3\,001$ anomalous labels in total
($\approx 1\%$ positives), with per-KPI event counts between $1$ and
$35$.

\subsection*{Evaluation}
We compare \textbf{C-MTAD-GAT} against two strong baselines on the TELCO
dataset: \textbf{MTAD-GAT} and \textbf{DC-VAE}. We also include a
\textbf{$\beta$-MTAD-GAT} variant that down-weights the KL term in the VAE loss (tuned $\beta{=}0.2$). In contrast to MTAD-GAT, C-MTAD-GAT uses a deterministic GRU reconstruction head and injects context via conditional convolutions, while DC-VAE is a Telco-specific dilated-convolutional VAE and the strongest published TELCO baseline. All models are trained on the same TELCO splits and use the \emph{same} unsupervised thresholding rule
(per-feature mean-based thresholds on validation errors), with no tuning on validation labels.

\textit{Pointwise timestamp-level.}
For each KPI we compute precision (P), recall (R) and F1 from the
timestamp binary anomaly labels. Because TELCO is strongly imbalanced, we also
report a prevalence-matched random baseline (\textbf{Random}), which
flags each timestamp as anomalous with probability equal to that KPI's
empirical anomaly prevalence, providing a natural performance floor.

\textit{Event-wise affiliation.}
Following \cite{huet2022local}, consecutive anomalous timestamps are merged into \emph{events}. Each predicted
event is matched to the ground-truth event it overlaps most; no overlap
counts as a false positive, unmatched ground-truth events as false
negatives. This yields event-level P, R and F1 that reward hitting
anomalous \emph{windows} rather than isolated timestamps.

\textit{Aggregation Macro, Micro and Union.}
We report \emph{Macro} (unweighted average over KPIs), \emph{Micro}
(P/R/F1 computed after pooling all predictions across KPIs)
and \emph{Union} (OR across KPIs per timestamp) variants of P, R and F1.

\begin{table}[t]
\centering
\footnotesize
\setlength{\tabcolsep}{3pt}
\renewcommand{\arraystretch}{1.1}
\begin{tabular}{l|ccc|ccc|ccc}
\toprule
& \multicolumn{3}{c|}{\textbf{Macro}} & \multicolumn{3}{c|}{\textbf{Micro}} & \multicolumn{3}{c}{\textbf{Union}} \\
\cmidrule(lr){2-4}\cmidrule(lr){5-7}\cmidrule(lr){8-10}
\textbf{Model} & P & R & F1 & P & R & F1 & P & R & F1 \\
\midrule
DC-VAE      & 0.644 & \best{0.682} & \second{0.663} & 0.694 & \best{0.589} & \second{0.637} & 0.561 & \best{0.798} & \second{0.659} \\
MTAD-GAT  & 0.589 & 0.339 & 0.430 & 0.412 & 0.201 & 0.270 & \best{0.601} & 0.446 & 0.512 \\
$\beta$-MTAD-GAT   & \best{0.716} & 0.591 & 0.647 & \best{0.793} & 0.420 & 0.549 & 0.549 & 0.674 & 0.605 \\
\midrule
\textbf{C-MTAD-GAT} & \second{0.707} & \second{0.673} & \best{0.690} & \second{0.781} & \second{0.544} & \best{0.641} & \second{0.582} & \second{0.776} & \best{0.665} \\
\bottomrule
\end{tabular}

\captionsetup{font=footnotesize}
\caption{TELCO — \textbf{Affiliation} scores. Best is \textbf{bold}; second-best is \second{shaded}. \textbf{Event Counts}. \emph{Micro}: Ground Truth (GT)=$143$, MTAD-GAT=$149$, $\beta$-MTAD-GAT=$429$, DC-VAE=$713$, \textbf{C-MTAD-GAT}=$503$. \emph{Union}: GT=$70$, MTAD-GAT=$133$, $\beta$-MTAD-GAT=$337$, DC-VAE=$567$, \textbf{C-MTAD-GAT}=$389$.}
\label{tab:telco_affiliation_agg_incell}
\end{table}

\begin{table}[t]
\centering
\footnotesize
\setlength{\tabcolsep}{3pt}
\renewcommand{\arraystretch}{1.1}
\begin{tabular}{l|ccc|ccc|ccc}
\toprule
& \multicolumn{3}{c|}{\textbf{Macro}} & \multicolumn{3}{c|}{\textbf{Micro}} & \multicolumn{3}{c}{\textbf{Union}} \\
\cmidrule(lr){2-4}\cmidrule(lr){5-7}\cmidrule(lr){8-10}
\textbf{Model} & P & R & F1 & P & R & F1 & P & R & F1 \\
\midrule 
Random  & 0.010 & 0.010 & 0.010 & 0.015 & 0.015 & 0.015 & 0.047 & \second{0.113} & 0.067 \\ \hdashline 
DC-VAE              & 0.220 & \second{0.096} & \second{0.134} & \second{0.127} & \best{0.063} & \best{0.084} & 0.103 & 0.110 & \second{0.107} \\
MTAD-GAT    & \second{0.350} & 0.028 & 0.052 & 0.078 & 0.012 & 0.021 & \second{0.112} & 0.042 & 0.061 \\
$\beta$-MTAD-GAT     & \best{0.385} & 0.071 & 0.120 & 0.111 & 0.045 & 0.064 & 0.097 & 0.081 & 0.088 \\
\midrule
\textbf{C-MTAD-GAT}      & 0.319 & \best{0.107} & \best{0.160} & \best{0.133} & \second{0.062} & \best{0.084} & \best{0.117} & \best{0.115} & \best{0.116} \\
\bottomrule
\end{tabular}
\caption{TELCO — \textbf{Pointwise} scores. Best is \textbf{bold}; second-best is \second{shaded}. 
\textbf{Timestamp Counts}.
\emph{Micro}: Ground Truth (GT)$=3001$, Random=3001, DC-VAE=1482, MTAD-GAT=476, $\beta$-MTAD-GAT=1209, \textbf{C-MTAD-GAT}=1389. 
\emph{Union}: GT$=1186$, Random=2850, DC-VAE=1268, MTAD-GAT=445, $\beta$-MTAD-GAT=991, \textbf{C-MTAD-GAT}=1163.}
\label{tab:telco_pointwise_agg_incell}
\end{table}

\subsection*{Results and Discussion}

We first examine \textit{event-level affiliation scores} in
Table~\ref{tab:telco_affiliation_agg_incell}, which measure how well
models hit incident \emph{windows}. The metric is forgiving: a trivial detector that flags every 
timestamp as anomalous has a minimum per-feature F1 score of $0.67-0.68$. 
Affiliation scores should therefore be interpreted together with the stricter pointwise
metrics to avoid over-crediting overly active detectors.

Across \textit{Macro/Micro/Union} aggregations, C-MTAD-GAT attains the highest
affiliation F1, with DC-VAE consistently second. $\beta$-MTAD-GAT reaches the best Macro/Micro
precision but at considerably lower recall, while MTAD-GAT is even more
conservative and misses many incidents. C-MTAD-GAT strikes a more balanced
precision–recall profile: recall is close to DC-VAE’s but with higher
precision, yielding better F1 and fewer alarms. On the union stream, our model
predicts substantially fewer incident windows than DC-VAE while maintaining a
higher F1, reducing the number of alarms that operators must inspect in daily operations. 
\textit{Pointwise} results in Table~\ref{tab:telco_pointwise_agg_incell} are much
smaller in absolute value because of severe class imbalance, but they provide a
useful counterweight to the affiliation metric. C-MTAD-GAT achieves the best
Macro and Union F1 and matches DC-VAE on Micro F1, while firing fewer positives
than both DC-VAE and the prevalence-matched random baseline. The random row
shows that even modest-looking F1 values correspond to clear gains over
chance. 

On the \textit{backbone} side, the VAE-based MTAD-GAT variants require careful
tuning of the KL weight and additional regularisation tricks and still
underperform the simpler GRU-AE head in C-MTAD-GAT under the same
unsupervised calibration. Standalone VAE variants (without MTAD-GAT)
showed similar sensitivity: performance degraded markedly when moving
away from hand-tuned settings, which is problematic in monitoring
pipelines that must be retrained regularly under resource and reliability
constraints. The deterministic GRU-AE configuration was more robust:
even non–fully-optimised hyperparameters yielded competitive performance,
which is crucial in large-scale operations.

Finally, on large national RAN and Core datasets with thousands of NEs
and hundreds of KPIs (not shown here), the same C-MTAD-GAT backbone converged faster and achieved higher
detection performance when we included static and dynamic context features. This motivates keeping context in the
architecture even though its effect on the single-NE TELCO benchmark is modest.

\section{Conclusion}

We proposed C-MTAD-GAT, a context-aware, centralised extension of
MTAD-GAT for unsupervised AD on mobile-network KPIs, and showed that on
the public TELCO dataset it outperforms MTAD-GAT and DC-VAE under a
shared label-free calibration protocol while emitting fewer alarms.
TELCO is a useful but simplified benchmark and only partially reflects
the scale and heterogeneity of RAN and Core deployments that motivated
our design. Although we focus on mobile networks, the same centralised, context-aware design applies to other industrial monitoring settings, such as large sensor networks in factories or power grids, where many similar units operate under different local conditions. In production, distribution shifts are currently handled by
periodic retraining; an important direction for future work is to make
C-MTAD-GAT explicitly shift-aware, for instance via residual-distribution
monitoring and adaptive thresholds. More principled uncertainty
quantification, beyond the current multi-seed approximation, could also
distinguish high- from low-confidence alarms and reduce costly false
positives, strengthening the resilience of AD in industrial operations.

\footnotesize
\bibliographystyle{unsrt}  
\bibliography{references} 

@inproceedings{mtad-gat,
  title={{Multivariate time-series anomaly detection via graph attention network}},
  author={Zhao, Hang and Wang, Yujing and Duan, Juanyong and Huang, Congrui and Cao, Defu and Tong, Yunhai and Xu, Bixiong and Bai, Jing and Tong, Jie and Zhang, Qi},
  booktitle={2020 IEEE international conference on data mining (ICDM)},
  pages={841--850},
  year={2020},
  organization={IEEE}
}

@article{garcia2023onemodel,
  title   = {One Model to Find Them All: Deep Learning for Multivariate Time-Series Anomaly Detection in Mobile Network Data},
  author  = {Garc{\'\i}a Gonzalez, Gast{\'o}n and Martinez Tagliafico, Santiago and Fern{\'a}ndez, Alejandro and G{\'o}mez, Gabriel and Acu{\~n}a, Juan and Casas, Pedro},
  journal = {IEEE Transactions on Network and Service Management},
  year    = {2023},
  doi     = {10.1109/TNSM.2023.3340146}
}

@article{Hundman_2018,
   title={Detecting {S}pacecraft {A}nomalies Using {LSTM}s and {N}onparametric {D}ynamic {T}hresholding},
   ISBN={9781450355520},
   url={http://dx.doi.org/10.1145/3219819.3219845},
   DOI={10.1145/3219819.3219845},
   journal={Proceedings of the 24th ACM SIGKDD International Conference on Knowledge Discovery \& Data Mining},
   publisher={ACM},
   author={Hundman, Kyle and Constantinou, Valentino and Laporte, Christopher and Colwell, Ian and S{\"{o}}derstr{\"{o}}m Tom},
   year={2018},
   month={Jul}
}

@inproceedings{DAGMM,
    title={Deep {A}utoencoding {G}aussian {M}ixture {M}odel for {U}nsupervised {A}nomaly {D}etection},
    author={Bo Zong and Qi Song and Martin Renqiang Min and Wei Cheng and Cristian Lumezanu and Daeki Cho and Haifeng Chen},
    booktitle={International Conference on Learning Representations},
    year={2018},
    url={https://openreview.net/forum?id=BJJLHbb0-},
}

@inproceedings{OmniAnomaly,
    author = {Su, Ya and Zhao, Youjian and Niu, Chenhao and Liu, Rong and Sun, Wei and Pei, Dan},
    title = {Robust {A}nomaly {D}etection for {M}ultivariate {T}ime {S}eries through {S}tochastic {R}ecurrent {N}eural {N}etwork},
    year = {2019},
    isbn = {9781450362016},
    publisher = {Association for Computing Machinery},
    address = {New York, NY, USA},
    url = {https://doi.org/10.1145/3292500.3330672},
    doi = {10.1145/3292500.3330672},
    booktitle = {Proceedings of the 25th ACM SIGKDD International Conference on Knowledge Discovery \& amp; Data Mining},
    pages = {2828-2837},
    numpages = {10},
    location = {Anchorage, AK, USA},
    series = {KDD '19}
}

@inproceedings{mad-gan,
  title={{MAD-GAN: Multivariate anomaly detection for time series data with generative adversarial networks}},
  author={Li, Dan and Chen, Dacheng and Jin, Baihong and Shi, Lei and Goh, Jonathan and Ng, See-Kiong},
  booktitle={International conference on artificial neural networks},
  pages={703--716},
  year={2019},
  organization={Springer}
}

@article{ramirez2023explainable,
  title={Explainable machine learning for performance anomaly detection and classification in mobile networks},
  author={Ram{\'\i}rez, Juan M and D{\'\i}ez, Fernando and Rojo, Pablo and Mancuso, Vincenzo and Fern{\'a}ndez-Anta, Antonio},
  journal={Computer Communications},
  volume={200},
  pages={113--131},
  year={2023},
  publisher={Elsevier}
}

@article{oleiwi2022mlts,
  title={{MLT}s-{ADCN}s: {M}achine learning techniques for anomaly detection in communication networks},
  author={Oleiwi, Haider W and Mhawi, Doaa N and Al-Raweshidy, Hamed},
  journal={IEEE Access},
  volume={10},
  pages={91006--91017},
  year={2022},
  publisher={IEEE}
}

@article{zheng2023transks,
  title={Trans{KS}: An {A}nomaly {D}etection {M}ethod for {T}elecommunication {N}etworks {B}ased on {D}eep {L}earning},
  author={Zheng, Jiahuan and Feng, Dongdong and Yang, Zhiming and Xiang, Yong and Zhang, Haiping and Li, Siyao},
  journal={IEEE Access},
  volume={11},
  pages={118048--118060},
  year={2023},
  publisher={IEEE}
}

@inproceedings{huet2022local,
  title={Local evaluation of time series anomaly detection algorithms},
  author={Huet, Alexis and Navarro, Jose Manuel and Rossi, Dario},
  booktitle={Proceedings of the 28th ACM SIGKDD Conference on Knowledge Discovery and Data Mining},
  pages={635--645},
  year={2022}
}

@inproceedings{
brody2022how,
title={How Attentive are {G}raph {A}ttention {N}etworks? },
author={Shaked Brody and Uri Alon and Eran Yahav},
booktitle={International Conference on Learning Representations},
year={2022},
url={https://openreview.net/forum?id=F72ximsx7C1}
}

\end{document}